\documentclass[runningheads]{llncs}
\usepackage{graphicx}
\usepackage{multirow}
\begin{document}
\title{About Evaluation of F1 Score for RECENT Relation Extraction System}
%
%
\author{Michał Olek\orcidID{0000-0003-2765-2570} }
\authorrunning{M. Olek}
%
\institute{Wrocław University of Science and Technology, Wrocław, Poland
\email{michal.olek@pwr.edu.pl}}
\maketitle              
\begin{abstract}
This document contains a discussion of the F1 score evaluation used in the article "Relation Classification with Entity Type Restriction" by Shengfei Lyu, Huanhuan Chen published on \textit{Findings of the Association for Computational Linguistics}: ACL-IJCNLP 2021. The authors created a system named RECENT and claim it achieves (then) a new state-of-the-art result 75.2 (previous 74.8) on the TACRED dataset, while after correcting errors and reevaluation the final result is 65.16 


\end{abstract}
\section{Introduction}

Relation extraction is the main topic of the article "Relation Classification with Entity Type Restriction" by Shengfei Lyu, Huanhuan Chen ~\cite{findings-2021-findings-recent} . The authors describe a system named RECENT that deals with extracting relations from sentences in documents where mentions (named entities) are already annotated. The code is available at \url{https://github.com//Saintfe/RECENT}. Technically, this kind of relation extraction task checks in the document each possible triplet (s, e1, e2) - where s is a sentence and e1 and e2 are mentions from the sentence - and assigns to the triplet a relation chosen from a set of predefined relations (set includes also special "\textit{no\_relation}” relation indicating that there is no relation between mentions). Often e1 mention is called subject and e2 mention is called object, as relations are, in general, directed.

Recent works focus mainly on making use of neural networks as in \cite{improved_baseline} or using various additional enhancements like curriculum learning  \cite{improving_curriculum_learn}. There are also approaches using two-way span prediction \cite{span-prediction} or graph convolution networks over pruned dependency trees  \cite{zhang-etal-2018-graph}.

\section{Dataset}
The dataset used is TACRED \cite{zhang2017tacred} - one of the most popular datasets for sentence-level relation extraction. In TACRED dataset, the mentions are not only just marked but also
their type is annotated. The subject mention is of type ORGANIZATION or PERSON and the
object mention is categorized as one of 16 types such as LOCATION, ORGANIZATION,
PERSON, DATE, etc. The names of relations contain also an abbreviation of type of subject mention - relations names
look like \textit{'org:city\_of\_headquarters', 'per:age', ‘per:date\_of\_birth’}

\section{Entity Type Restriction}
The authors noticed that it is very often the case that a type of relation determines the types of
possible mentions involved and vice versa. So, for example, if we have a relation  \textit{per:date\_of\_birth} then it expects the subject to be a PERSON and the object to be a DATE.
Same the other way - if we have another sentence where the type of subject is ORGANIZATION and the type of object is CITY, one does not need to perform any additional checking to know that relation \textit{per:date\_of\_birth} is not a valid relation here. But relation \textit{org:city-of-headquarters} may be.

So we can regroup the data into the independent subsets, where each subset contains sentences with just one pair of types of subject and object i.e. (ORGANIZATION,PERSON). And then for each such subset, we can construct a set of all matching relations types from relations that can be found in at least one sentence of the subset. The list would look like  ( \textit{no\_relation, org:founded\_by, org:shareholders, org:employees}). The number of relation types is usually significantly smaller for the subset than for the whole set. With such regrouping, we can decompose the whole task of relations extraction to many similar smaller independent tasks and train a specific, separate  \textit{semantic classifier} for each such subset. Since, generally, the number of relation types is smaller for each subset than for the whole set, we assume that for each subset the independent semantic classifier - focused only on that particular subset -  should perform better than the classifier trained on the whole set. So we process each subset using a corresponding semantic classifier and after gathering all the results from all semantic classifiers we should get better overall result for the whole data set. Since this approach is model-agnostic the system RECENT was tested with two models: CGN \cite{zhang-etal-2018-graph} and SpanBERT \cite{10.1162/tacl_a_00300}. 
For the first model reported result was 70.9 F1 score, and for the second model 75.2 F1 score.

\section{Implementation}

In the beginning, the whole dataset is divided into the aforementioned subsets. Normally it would be divided into quite a few subsets because there are quite a few combinations that can be made of 2 subject mention types and 16 object mention types. But authors noticed that all except 13 subsets are so simple that for each such simple subset any triplet can be assigned just to either one relation specific to this subset or to "\textit{no\_relation}". 

\subsection{Binary classifier}

For example, in the subset constructed from subject mention type PERSON and object mention type RELIGION the only meaningful relation that can be found  between mentions in training data is \textit{per:religion}. In another subset, where the subject mention type is PERSON and object mention type is TITLE the only meaningful relation possible to be found in training data is \textit{per:title}. We infer from training data what is a relation that such subset may assign triplets to, so basically, the relation extraction task for these kinds of subsets is reduced just to decide whether there is a relation for triplet or whether there is not. It can be done with a \textit{binary classifier}. This is classifier trained to recognize just whether there is a meaningful relation in a given triplet and does not say what kind of relation it is. The classifier is trained once on the whole train dataset. During evaluation, this is the first step: the binary classifier checks if given triplet is "\textit{no\_relation}" and if it is true the final answer is  "\textit{no\_relation}". Second step checks if the triplet is a one from the simple subsets and if it is true then the final answer is a relation associated with this simple subset.

So the only ones left are triplets from the other, more complicated, subsets and they are to be dealt with the semantic classifiers.

\subsection{Semantic classifiers}

The semantic classifiers need to be trained only for the 13 more complicated subsets. For each such subset, there are three helper files generated, containing the corresponding train, tune, and test examples and one additional file with a list of possible relations that can be found in triplets of that subset. Each semantic classifier is trained to recognize which one of these possible relations is associated with a given triple from the particular subset. 

However, we need to keep in mind that the very first step of evaluation is binary classifier filtering out all triplets considered by it as associated with "no\_relation".

With this setup, the semantic classifiers do not deal with recognizing "\textit{no\_relation}". They have to assume that if something is passed to them it contains only  "meaningful" relation. This way they do not contain "\textit{no\_relation}" label in available answers. So, to train them well, \emph{all} helper files ( training, tune, and test) created for semantic classifiers are cleared of samples with the answer "\textit{no\_relation}". The resulting training and tuning files are used to train semantic classifiers. The resulting test files are used for the pre-evaluation step described in the next section

\hfill

Sidenote: each one of these additional files mentioned at the beginning of paragraph and containing the list of possible relations for the corresponding subset is generated using data files with \emph{test} examples but should be generated using files with train examples. However, the data is so homogeneous that correcting this issue has no significant effect on the final result.

\section{Evaluation}

\noindent
The whole idea of prediction for a sample goes as follows:     

\begin{enumerate}

\item Sample is checked against the binary classifier. If it says it is "\textit{no\_relation}" then this is the final answer for this sample.
\item Otherwise, sample is checked against simple subsets. If it is an element of such a subset then the final answer is the one relation associated with the subset. 
\item Otherwise, the corresponding complicated subset is identified and the sample is checked by the semantic classifier trained for this particular subset.

\end{enumerate}

\noindent
Actually, in the third step, the semantic classifiers are not used. Very likely to avoid a situation where all the models for semantic classifiers are present in the memory at the same time. Checking is actually done against the pre-computed \textit{partial final result file} for the corresponding subset. The file is made even before actual evaluation phase: during this pre-evaluation step, results are computed and the partial files are created - each semantic classifier is loaded into memory once and is run over the helper test file corresponding to the associated subset and the results are stored in \textit{"partial final result files"}, and then, in step 3, the values contained in files are used as final predictions for the more complicated subsets.

\noindent
So, in the third step the system checks the file if it contains the given triple:
\begin{itemize}
    \item 3.1 if the sample is found then its corresponding answer is the final result
    \item 3.2 otherwise the final answer is set to '\textit{\textit{no\_relation}}'
\end{itemize}

\noindent
The last point may come as a bit of a surprise. It is not mentioned in the article but it is how the code works (lines 56-59 in file SpanBERT/recent\_eval.py). Since the binary classifier allowed the sample to pass through, there should be a meaningful answer. How can it be that for samples classified by the binary classifier as the ones that should be associated with a relation, and which should be processed by semantic classifiers, the system is coded to give an answer as \textit{no\_relation}?

The solution is simple: the binary classifier is not perfect and there are cases when it classifies the sample as the one that should be associated with meaningful relation even if actually the sample is associated with \textit{no\_relation}

The semantic classifiers were taught on purpose to always give a meaningful relation as an answer and they are not capable to handle this situation. So it is hard-coded in the system that in such a situation the final answer should be \textit{no\_relation}. Is it correct, and if yes, why?

\subsection{Evaluation of a single sample}

Let us take for example a sample with id '098f665fb92fee9d29b3' - the last sample in the test data set. Its subject type is ORGANIZATION and its object type is PERSON and there is a semantic classifier trained for this pair of types. There is no relation in our sample but the binary classifier happened to incorrectly classify this sample as the one with a meaningful relation. And if we manually process step 3 we check against preprocessed \textit{partial final result file} for the corresponding semantic classifier  - the sample is not found there, so the system classifies the sample as having \textit{no\_relation}. Correctly, but only because the sample was not found in pre-processed helper test file. Why it is not found there? It is not there because during preprocessing - as described at the end of paragraph 4.2 - all samples with "\textit{no\_relation}" were filtered out from the \emph{test} data.

Normally, with fresh production data - let us assume we are processing a completely
new piece of data without test dataset ( when there is no pre-processed
test data provided) - the sample like the one described above would be classified
incorrectly. It would be classified as associated with one of the meaningful
relation types associated to corresponding semantic classifier since the classifier
is unable to yield ”no relation”. as a final result. And it ”does not know”
that binary classifier was wrong. Simply, if test data were not provided earlier
then there will be no filtering out the samples with ”no relation”.

To put it short: the system says that this particular sample from test data should be classified as ”no relation” because the system checked earlier that in the test data it is classified as ”no relation”. It happens to all samples that belong to complicated subsets, have no meaningful relation, and are incorrectly classified by the binary classifier as the ones which have meaningful relation.

All these samples are classified correctly due to this loophole. After closing this loophole all these samples are classified incorrectly as false positives since the semantic classifiers are unable to yield the answer "no\_relation".

\section{Reevaluation}

To calculate the actual value of F1 a new experiment is needed with such a modification that corrects the issue described in the previous paragraph. All the used models are exactly the same as in the original experiment but in cases where workflow comes to point 3, a sample is always given to the corresponding semantic classifier to be classified - without checking if the sample is present in the appropriate "partial final result file”. In terms of calculating the F1 score, and technical realization, the above setup is equivalent to just changing step 3.2 to yield any meaningful relation except the correct one: "no\_relation", since the semantic classifiers are unable to return "no\_relation". The task is multiclass classification where one does not count in true positives for "no\_relation".

Table 1 presents the results of recreated original experiment and the results of experiment with correction closing the loophole. One can see that only the number of false positives has changed.

\begin{table}[ht]
\caption{Comparison of results of experiments}
\begin{center}

\begin{tabular}{c c c c }
    \hline    
    & True Positive (TP)   \phantom{.}&  False Positive (FP)  \phantom{.}& False Negative (FN)  \phantom{.} \\ 
    \hline
    Recreated original experiment & 2182 & 246 & 1143 \\ 
    \hline
    Experiment with correction & 2182 & 1190 & 1143 \\ 
    \hline
\end{tabular}
\end{center}
\end{table}

\noindent
To calculate F1 score from these results we need calculate precision and recall.
Precision is calculated using formula:

\begin{equation}
P = \frac{TP} {TP + FP}
\end{equation}
        
\noindent
and recall is calculated using formula:

\begin{equation}
R = \frac{TP}{TP + FN}
\end{equation}

\noindent
and F1 score is calculated as a harmonic mean of \emph{P} and \emph{R}:

\begin{equation}
F1 = \frac {2 * P * R} {P + R}
\end{equation}

\begin{table}[ht]
\caption{Comparison of precision, recall and F1 (x100)}
\begin{center}
\begin{tabular}{c c c c}
    \hline    
    & Precision    \phantom{.}& Recall    \phantom{.}&\phantom{.}    F1   \phantom{.}  \\ 
    \hline
    Recreated original experiment & 89.86 & 65.62 & 75.85 \\ 
    \hline
    Experiment with correction & 64.70 & 65.62 & 65.16 \\ 
    \hline
\end{tabular}
\end{center}
\end{table}

Precision, recall, and F1 score are presented for both experiments in table 2.
One can see that recall is the same in both experiments since only the number of false positives has changed.

After calculating the actual value for the F1 score for system RECENT using SpanBERT model is 65.16, which is 10 pp. lower than originally reported.

When evaluating using CGN model the drop in precision is the same and the reevaluated F1 score for this model is 61.

The idea was very clever and promising but the results show that it does not give the expected improvement in performance. Perhaps this is due to the fact that each individual classifier learns on a much smaller dataset than such a single general one. And the loss from having less data to learn is greater than the gain gained from allowing the classifier to focus on only a small subset of the responses.

%
%
%
%

\end{document}